\documentclass[letterpaper, 10 pt, conference]{ieeeconf}

\makeatletter
\IEEEtriggercmd{\reset@font\normalfont\scriptsize}
\makeatother
\IEEEtriggeratref{1}

\usepackage{subcaption}

\DeclareCaptionLabelSeparator{periodspace}{.\quad}
\makeatletter
\let\NAT@parse\undefined
\makeatother

\usepackage[dvipsnames]{xcolor}
\usepackage[normalem]{ulem}

\usepackage[ruled,noend,linesnumbered]{algorithm2e}
\usepackage{float}
\usepackage[pagebackref=true,breaklinks=true,colorlinks,bookmarks=false]{hyperref}

\usepackage{tikz}
\usepackage{textcomp}
\usepackage{lipsum}

\usepackage{bbm}
\usepackage{amssymb}
\usepackage{amsmath}
\usepackage{multirow}

\usepackage[capitalise]{cleveref}

\usepackage{wrapfig}

\renewcommand{\baselinestretch}{0.96}

\newcommand{\ignore}[1]{}

\IEEEoverridecommandlockouts                           
\title{\LARGE \bf
Towards Autonomous Grading In The Real World
}

\author{Yakov Miron$^{1,2}$, Chana Ross$^{1}$, Yuval Goldfracht$^{1}$, Chen Tessler$^{1}$ and Dotan Di Castro$^{1}$ 
\thanks{{\tt \{yakov.miron, chana.Ross, yuval.goldfracht, dotan.dicastro\}@il.bosch.com, chen.tessler@gmail.com}.
}
\thanks{$^1$Bosch Center for Artificial Intelligence (BCAI), Haifa, Israel}
\thanks{$^2$The Autonomous Navigation and Sensor Fusion Lab, The Hatter Department of Marine Technologies, University of Haifa, Israel} 
}

\date{2022}

\begin{document}

\makeatletter
\g@addto@macro\@maketitle{
  \begin{figure}[H]
  \setlength{\linewidth}{\textwidth}
  \setlength{\hsize}{\textwidth}
    \centering

  \includegraphics[width=0.99\linewidth, height=8cm]{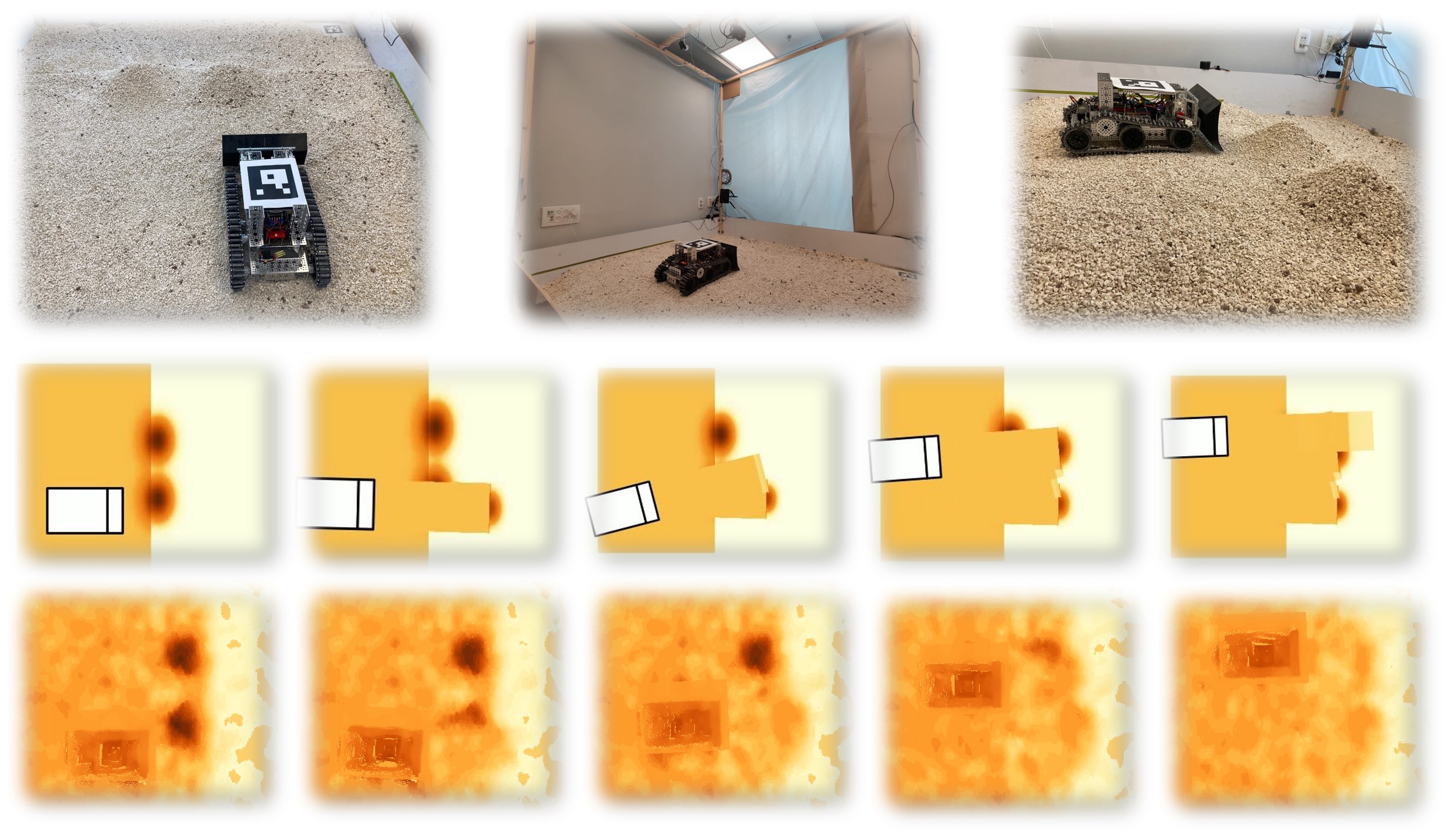}
  
  \caption{A comparison of grading policy between a simulated bulldozer and a real-world scaled prototype on an area containing sand piles. Here, the agent is provided with an initial graded area and is required to extend it by pushing the sand piles.
  \textbf{Top Row:} Photos of our experimental setup (see \cref{subsec: prototype dozer}) showing the scaled bulldozer prototype facing the sand piles. 
  \textbf{Middle {\&} Bottom Rows:} Heightmaps extracted from our simulation and real-world scaled setup respectively. Sand piles are captured as dark blobs that indicate their height. In each column, we compare the grading policy in simulation to the real-world setup, on a similar scenario.
  }
    \label{fig:sim2real_all} 
  \end{figure}
  }

\thispagestyle{empty}
\pagestyle{empty}

\maketitle

\setcounter{figure}{1}

\begin{abstract}

Surface grading is an integral part of the construction pipeline. Here, a bulldozer, which is a key machinery tool at any construction site, is required to level an uneven area containing pre-dumped sand piles. In this work, we aim to tackle the problem of autonomous surface grading on real-world scenarios. We design both a realistic physical simulation and a scaled real-world prototype environment mimicking real bulldozer dynamics and sensory information. In addition, we establish heuristics and learning strategies in order to solve the problem. Through extensive experiments, we show that although heuristics are capable of tackling the problem in a clean and noise-free simulated environment, they fail catastrophically when facing real-world scenarios. However, we show that the simulation can be leveraged to guide a learning agent, which can generalize and solve the task both in simulation and in a scaled prototype environment.

\end{abstract}

\section{Introduction} \label{Introduction}

Recent years have seen a dramatic rise in demand for automation in the construction industry. Bulldozers play a key role at most construction sites and are the go-to machinery for many tasks. While hand-operated bulldozing is a well established profession, it suffers from a shortage in experienced drivers, which leads to longer construction times and higher costs. Our work focuses on automation of the grading task, where a bulldozer is confronted with an uneven plot of land (see \cref{fig:sim2real_all}). The bulldozer is equipped with a blade and pushes around the soil to bring the plot to a specified height. 

Similarly to other real-life robotics applications, surface grading suffers from two major challenges. First, missing or noisy sensory information, and the agent's limited field-of-view makes the problem partially observable. This makes autonomy difficult compared to a fully observable scenario. Second, data-driven methods e.g., deep neural networks (DNNs), are sample inefficient and require large amounts of expert data in order to converge. In the context of surface grading, this aspect is especially challenging as obtaining diverse real-world data is expensive and sometimes infeasible. Specifically, it requires long hours of manual operation, of multiple machines, in order to design the grading scenario i.e., clearing the work area, placing sand piles, calibrating equipment, etc.

In order to learn a behavior policy capable of autonomous grading in the real-world, we design a simulator and leverage it using deep supervised learning techniques. As opposed to the real-world, a good simulator provides several benefits. First, data collection does not require an expert human operator and can instead be obtained by running advanced heuristics on clean, privileged data. Second, simulations are configurable and therefore enable the design of complex scenarios and data augmentation techniques. Moreover, simulations enable the domain expert to focus on the challenging parts of the task in hand. Finally, by using a simulator, we can collect data for training and also efficiently evaluate a policy  without the risk of failure in the real-world. Therefore, reducing the risk and wear-and-tear on real bulldozers.

In this work, we focus on DNNs for perception and decision making. Unlike classic rule-based methods, DNNs are trained to make decisions based on collected data. While classic detection techniques rely on the ability to detect edges, a DNN is trained to extract more complex features. By doing so, it can overcome data imperfections such as missing information due to occlusions and measurement noise. We utilize a method called privileged learning \cite{vapnik2009new}, where the agent is trained on noisy data to imitate an expert, who has direct access to perfect and noise-free measurements. We show how this paradigm learns a robust feature extractor, capable of overcoming the inaccuracies of real-world measurements. Moreover, our agent trained using privileged learning is robust and able to generalize, while methods that rely on classic detection techniques \cite{hirayama2019path} fail when applied to our real-world scaled prototype environment, 

Our main contributions are as follows: \\
\textbf{(1)} We create a physically realistic simulation environment for training and evaluation of both heuristics and learning-based agents.\\
\textbf{(2)} We show that training agents using privileged learning techniques can be used to overcome the limitations of standard methods for autonomous grading. \\
\textbf{(3)} We validate our methods and assumptions on a scaled prototype environment, which includes real-world vehicle dynamics, sensors and soil interaction.

\section{Related Work} \label{sec:related_work}

\subsection{Bulldozer Automation}\label{subsec:related_work_bulldozer_automation}

Very little research has been done on the automation of bulldozers, specifically, on path-planning optimization using deep learning methods. One explanation might be the complexity of this task. Specifically, manual operation of a bulldozer requires skillful operator as the behavior of the soil is irreversible \cite{6030650}. 

In \cite{hirayama2019path}, the authors implemented a heuristic approach for autonomous surface grading. They examined the trade-off between grading the pile when the blade is at full capacity and pushing less sand in order to reduce the elapsed time.
\cite{nakatani2018autonomous} and \cite{9385686} used off-the-shelf reinforcement learning (RL) techniques (DQN; \cite{mnih2015human}) on an overly simplified environment with a single sand pile. Their simulation did not include the change in the bulldozer's velocity as a function of volume being pushed, nor did they validate their assumptions in real-world experiments.

\subsection{Sand Simulation} \label{subsec:related_work_sand_simulation}

Precise particle simulation, in the case of sand, for example, is an active line of research using both classic and modern tools. Classical methods describe soil using solid mechanical equations \cite{solid_mechanics}, discrete element methods \cite{finite_elements}, or fluid mechanics \cite{SULSKY1994179}; whereas newer modern methods utilize DNNs \cite{sanchez2020learning} to simulate the reaction of particles to forces. While these methods achieve outstanding results and have impacted both the gaming and cinematic industries, they require high computational cost and long run-times. \cite{ross2021agpnet} established a simulator for earth-moving vehicles, where the goal was to enable policy evaluation for bulldozers. This simulation was fairly accurate, fast and able to capture the main aspects of the interaction between the sand and the vehicle. 

As our goal is to enable rapid data collection and policy evaluation, we chose to simulate the soil by considering only the heightmap \cite{developing_simple} i.e only the surface of the soil changes over time. This approach takes into account key interactions on the one hand while maintaining simplicity and efficiency on the other, and is commonly used for robotic applications.

\subsection{Sim-to-Real} \label{subsec:related_work_sim2real}
The task of training in simulation and deploying in the real-world has been of much interest in the past years and is known as \textit{sim-to-real}. It can be divided into two main categories -- dynamics and perception. 
The \textit{dynamics} gap stems from the inability to precisely model the reaction of the system. This is tackled by solving a robust objective using methods such as dynamics randomization \cite{peng2018sim} or adversarial training \cite{pinto2017robust,tessler2019action}.
Overcoming the \textit{perception} gap is done by learning a mapping between the simulation and the real-world \cite{rao2020rl}, \cite{miron2019sflow} or learning robust feature extractors \cite{loquercio2021learning}.

Comparing to previous work, we create a simulation that closely mimics  real dynamics, thus enabling us to focus our efforts on closing the visual \textit{sim-to-real} gap. 
We chose to minimize this gap by using a heightmap to represent the environment within our simulation. This is beneficial compared to RGB images as it better resembles the real-world data \cite{loquercio2021learning}. Finally, we leverage privileged information, available only during simulation, in order to train a policy capable of overcoming the inaccuracies of real-world data.

\section{Background and Problem Formulation}
We begin by formally describing the task of an autonomous bulldozer, the setup, imitation learning and privileged learning.

\label{sec:problem_formulation}

\subsection{Partially Observable Markov Decision Processes}\label{subsec: mdp}

A Partially-Observable Markov Decision Process (POMDP; \cite{sutton2018reinforcement}) consists of the tuple $(\mathcal{S}, \mathcal{O}, \mathcal{A}, \mathcal{P}, \mathcal{R})$. The state $s \in \mathcal{S}$ contains all the required information to learn an optimal policy. However, agents are often provided with an observation $o \in \mathcal{O}$, which contains partial or noisy information regarding the environment. Unlike states, observations typically lack the sufficient statistics for optimality. At each state $s \in \mathcal{S}$ the agent performs an action $a \in \mathcal{A}$. Then, the system transitions to the next state $s'$ based on a transition kernel $P(s' | s, a)$. Finally, the agent is provided with a reward $r(s, a)$, which scores the chosen action according to a predefined metric. 

The goal of an agent is to learn a behavior policy $\pi(s)$, which can be stochastic or deterministic, that maximizes the cumulative reward-to-go. In this work, we assume a POMDP/R i.e. a POMDP without a reward function.

\subsection{Problem Formulation}\label{subsec: grading as mdp}

\begin{figure}
 \centering
 \begin{subfigure}[b]{0.15\textwidth}
     \centering
     \includegraphics[width=\textwidth]{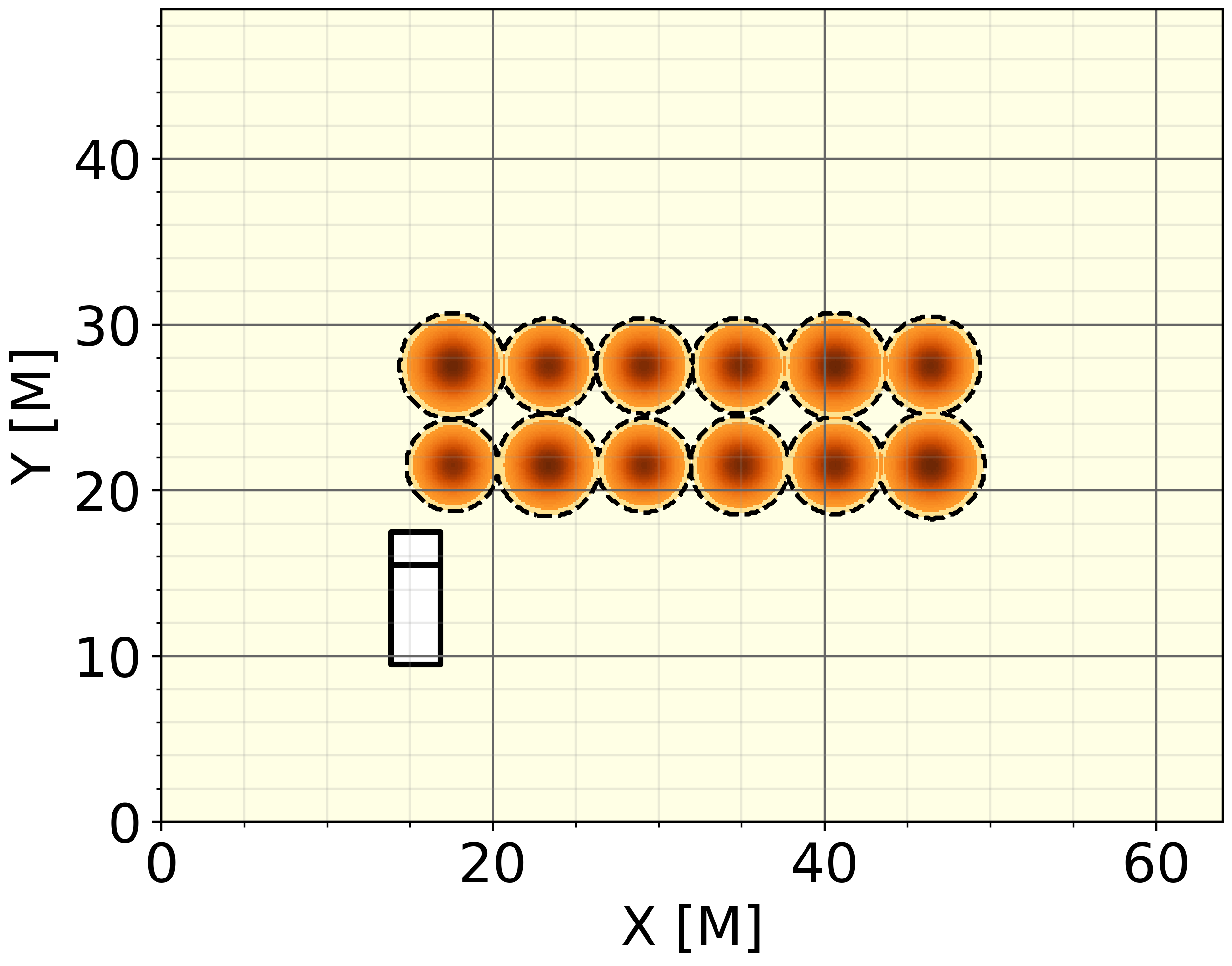}
     \caption{Initialization}
     \label{fig:scenario_init}
 \end{subfigure}
 \hfill
 \begin{subfigure}[b]{0.15\textwidth}
     \centering
     \includegraphics[width=\textwidth]{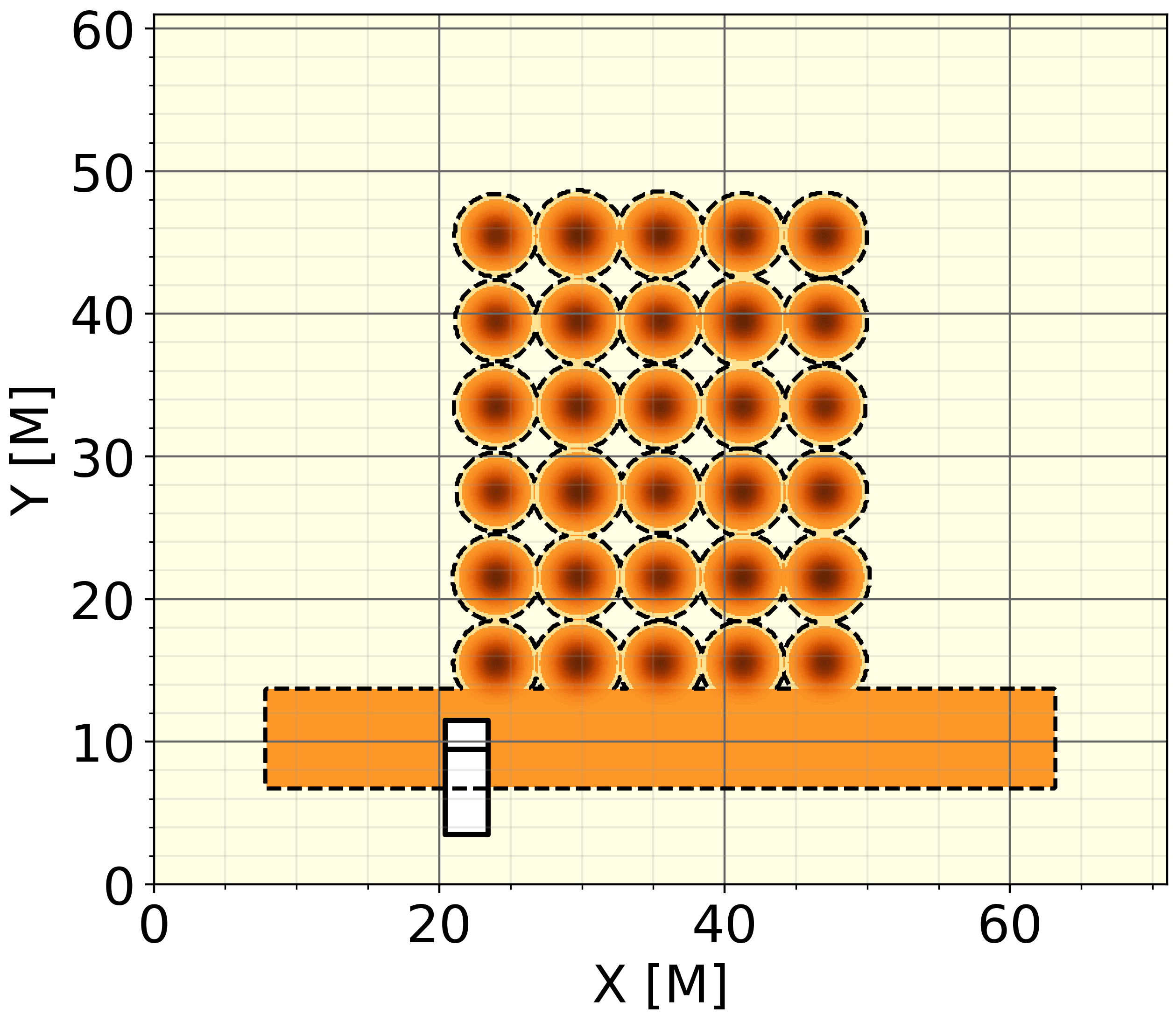}
     \caption{Continuous}
     \label{fig:scenario_cont}
 \end{subfigure}
 \hfill
 \begin{subfigure}[b]{0.15\textwidth}
     \centering
     \includegraphics[width=\textwidth,trim={0 0 0 4.2cm},clip]{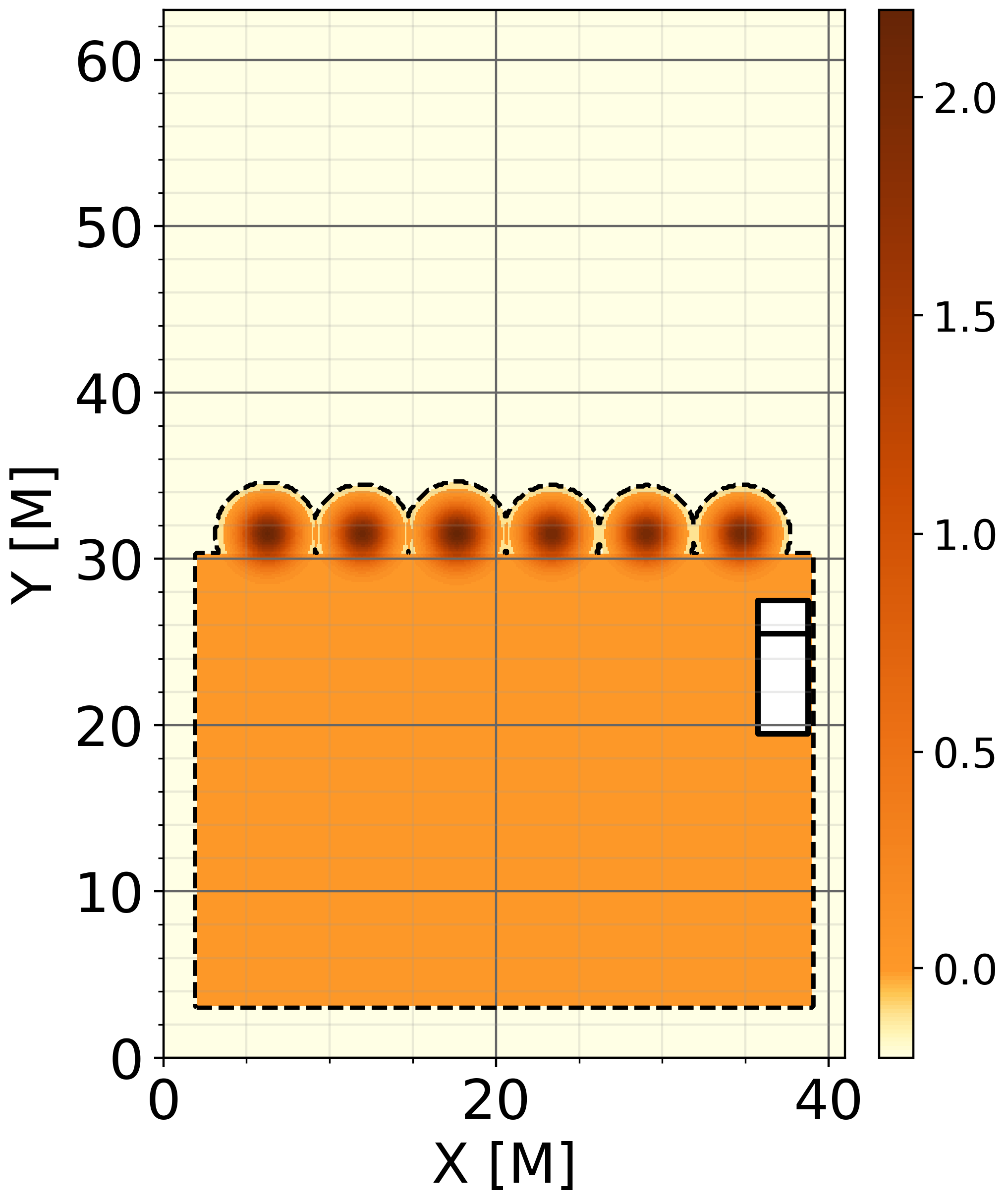}
     \caption{Edge}
     \label{fig:scenario_edge}
 \end{subfigure}
    \caption{The autonomous grading task is divided into three sub-tasks. \textbf{(a)} \textit{initialization}: Here, the area contains a few rows of pre-dumped sand piles without any previously graded area. The bulldozer is required to create an incline and reach a predefined target height. \textbf{(b)} \textit{continuous}: Here, the bulldozer is located on top of the previously graded platform, and sand piles are continuously being added to the vicinity of the graded platform. The bulldozer is required to extend the graded platform. \textbf{(c)} \textit{edge}: Here, most of the area is already graded and sand piles are dumped at the edge of the platform. The bulldozer is required to create decline from the platform in order to safely leave the site.}
    \label{fig:scenario_examples} 
\end{figure}

In order to tackle the task of autonomous grading, we divide it into three main sub-tasks: \textit{initialization}, \textit{continuous}, and  \textit{edge} (see \cref{fig:scenario_examples}). Each sub-task exhibits different challenges and can be simulated and tested individually. Next, we formalize it as a POMDP/R as described in \cref{subsec: mdp}.

\begin{figure}
 \centering
  \begin{subfigure}[b]{0.21\textwidth}
     \centering
     \includegraphics[width=\textwidth]{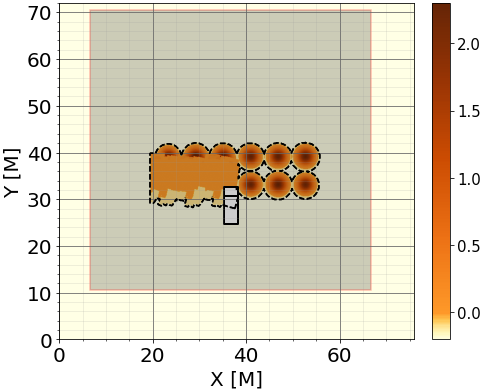}
     \caption{State}
     \label{fig:state}
 \end{subfigure}
 \hfill
 \begin{subfigure}[b]{0.22\textwidth}
     \centering
     \includegraphics[width=\textwidth]{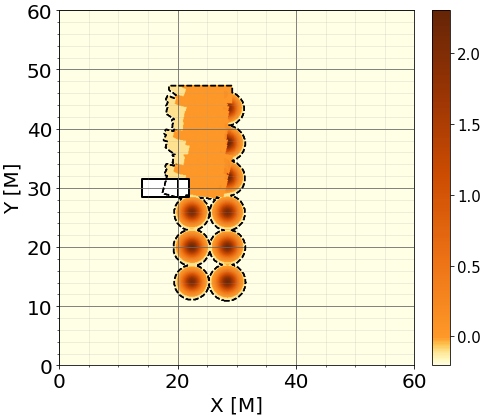}
     \caption{Observation}
     \label{fig:observation}
 \end{subfigure}
    \caption{\textbf{(a)} An example of the full state, which includes all of the information regarding the task (full heightmap of the area and bulldozer position). The projection of the observation to the coordinate system of the state is marked in gray \textbf{(b)} The state's corresponding observation, which is a bounding box view of the state around the bulldozer, described in the bulldozer's \textit{ego} axis.}
    \label{fig:state_obs} 
\end{figure}

\noindent \textbf{States:} In our case, the state includes  the target area size, the bulldozer's location within that area, the heightmap of the target area, and the full bulldozer trajectory up until the current time point. see \cref{fig:state} for additional details.

\noindent \textbf{Observations:} In our case, the observation is partial due to two reasons. First, the construction site's dimensions may vary while the agent's field-of-view remains constant. Second, measurement errors are common in real-world sensors and often lead to missing or noisy information. Therefore, the agent is presented with an  \textit{ego-view} of the current state. Meaning, a bounding box view around the current location of the bulldozer (marked in grey in \cref{fig:state}) together with some measurement noise augmentations that mimic real-world sensors (see \cref{fig:observation}).

\noindent \textbf{Actions:} The bulldozer's control can be performed both at the low level, e.g., providing rotation and velocity to the bulldozer, or at a higher level, e.g., selecting a destination coordinate. We chose to focus on macro-actions \cite{sutton1999between}, such as coordinate selection, and leave the low-level control to classic control algorithms. Macro actions, also known as skills or options, are temporally extended actions and are known to benefit learning especially when those are meaningful. Taking inspiration from the behavior of expert bulldozer operators, we consider the \textit{start-point and push-point} action-set ,abbreviated as `\textit{SnP}', inspired by \cite{hirayama2019path} and shown in \cref{fig:snp_example}. At each state, the agent selects two coordinates, \textit{push-point} and \textit{start-point} denoted as $(P, S)$, respectively. The bulldozer drives towards $P$, then reverses slightly further than its initial location to an intermediate point $B$, and finally moves to the next starting position $S$.

\noindent \textbf{Transitions:} Transitions are determined by the bulldozer's dynamics and physical properties of soil and the environment. 

\subsection{Imitation Learning}\label{subsec:imitation}

Imitation learning (IL; \cite{hussein2017imitation}) is the go-to approach for learning behavior policies in sequential decision making problems. As opposed to reinforcement learning (RL), it relies on expert demonstrations of the correct policy and is highly suitable for cases where the definition of a reward function does not naturally exist.  

In this work, we focus on behavior cloning (BC; \cite{ross2010efficient}), where an expert demonstrator collects data a-priori. This data is then used offline to train a behavior policy using a supervised learning loss: 
\begin{equation} \label{eq: imitation loss}
    L(\theta; D) = \underset{(o, a) \sim D }{\mathbb{E}} \Bigl[ l\bigl( \pi_\theta (o), a\bigr)\Bigr] ,
\end{equation}

Here, $D$ is the training dataset generated by an expert demonstrator containing corresponding actions $a$, and observations $o$. $\pi_{\theta}$ is the agent's learned policy and $l(\cdot)$ is a cross entropy metric.

While more advanced IL methods exist, such as DAgger \cite{ross2011reduction}, they generally require low-level expert interaction. For instance, an expert driver can demonstrate how to drive on the highway. Yet, given a single observation it is often impossible to provide the precise steering angle and throttle required. Surface grading is no different in that sense, making these methods inadequate for the task.




\subsection{Privileged Learning}\label{subsec: priv learning}

In classic machine learning (ML), the goal of a teacher (expert) is merely used to transfer knowledge to the student (agent). In many tasks, the expert can obtain access to additional information, during training, that is not available during inference. This technique is called privileged learning \cite{vapnik2009new}. The benefit of this scheme, is the ability to leverage the data-driven nature of machine learning models, allowing them to learn robust estimators.

\begin{figure}
    \centering
    \begin{subfigure}[b]{0.15\textwidth}
        \includegraphics[width=\textwidth]{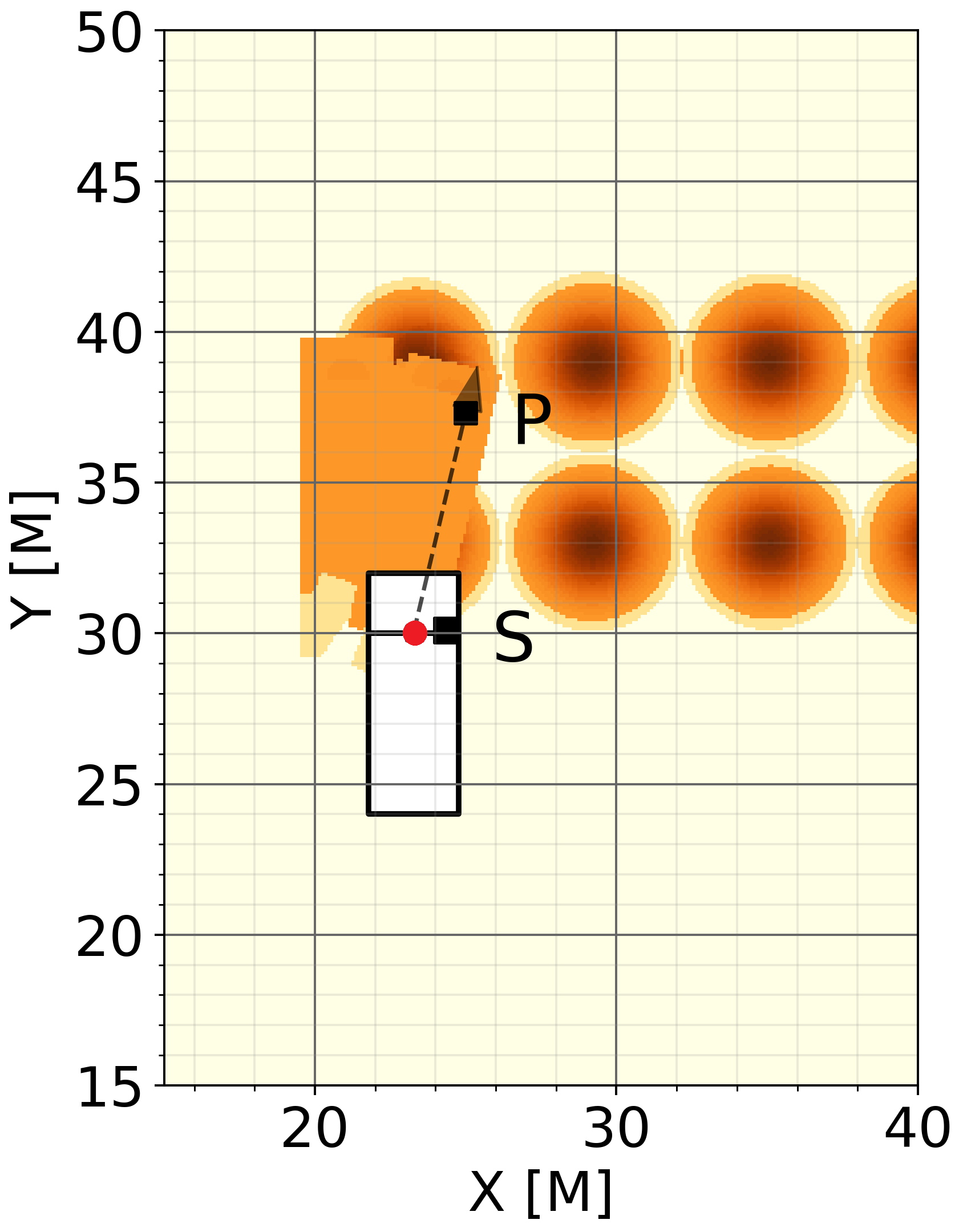}
        \caption{Initial state}
        \label{fig:snp_exam1}
    \end{subfigure}
    \begin{subfigure}[b]{0.15\textwidth}
        \includegraphics[width=\textwidth]{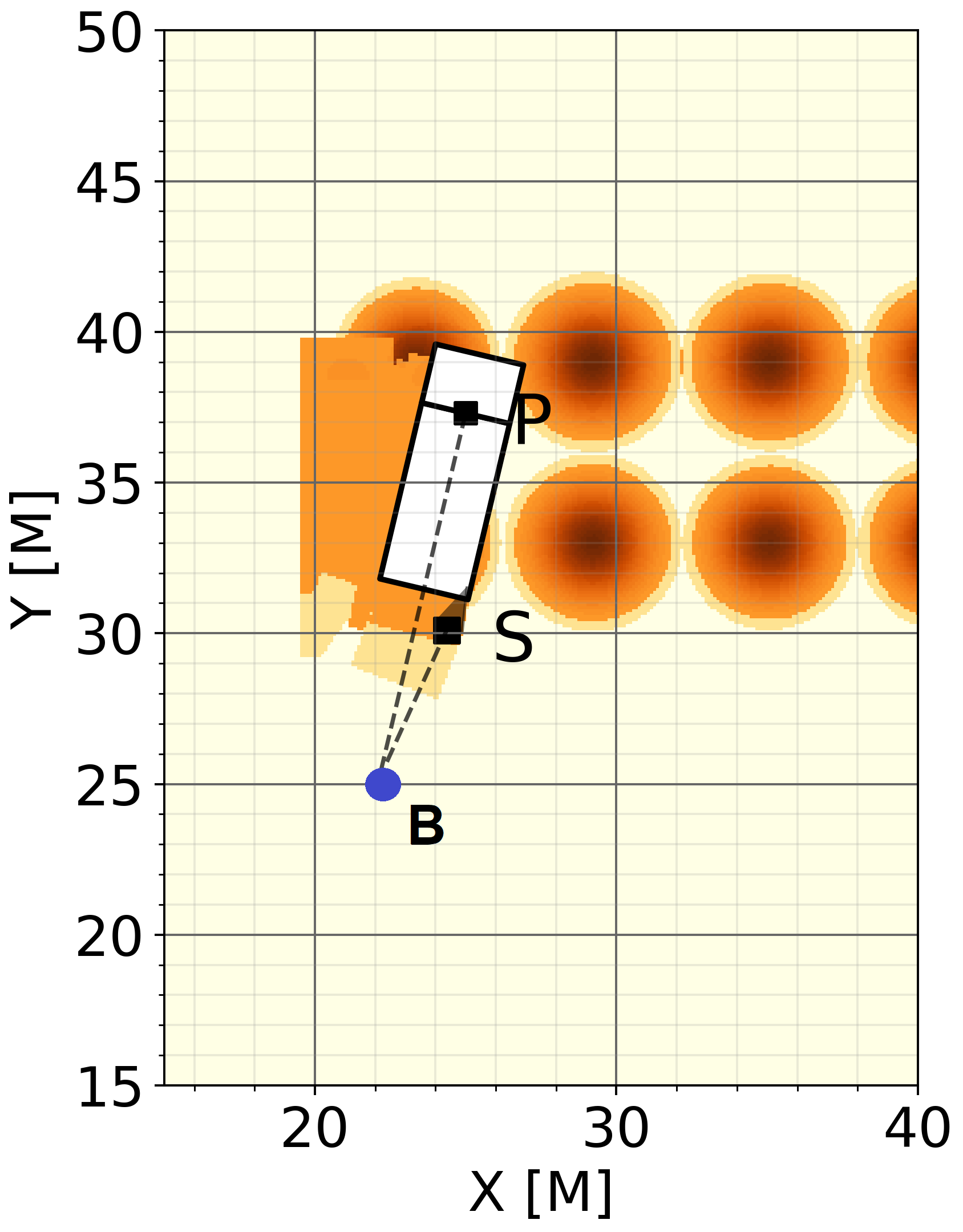}
        \caption{State after $p$}
        \label{fig:snp_exam2}
    \end{subfigure}
    \begin{subfigure}[b]{0.15\textwidth}
        \includegraphics[width=\textwidth]{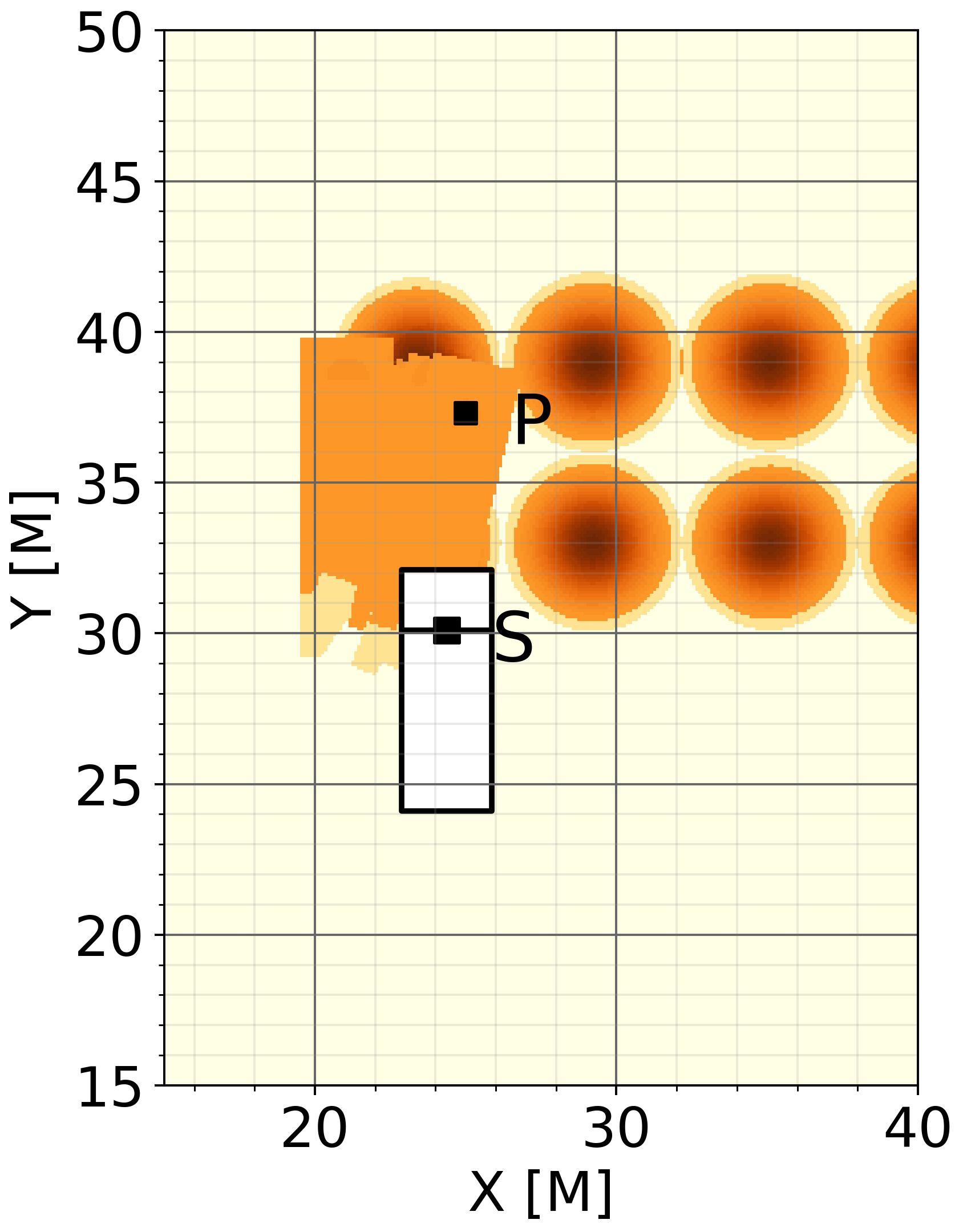}
        \caption{State after $s$}
        \label{fig:snp_exam3}
    \end{subfigure}
    \caption{Example of the trajectory, which the bulldozer follows when executing actions ($p$, $s$). The initial position is marked by the red dot in \textbf{(a)}. 
    The agent executes the following maneuver:
        (i) From origin rotate to face $p$.
        (ii) Drive forward to $p$. Only here, the bulldozer interacts with sand while grading it.
        (iii) Reverse back to B (blue dot in \textbf{(b)}).
        (iv) Rotate to face next $s$.
        (v) Drive forward to next $s$ in \textbf{(c)}.
    } 
    \label{fig:snp_example}
\end{figure}

\section{Method}\label{sec:method}


In order to tackle the problem of autonomous grading we focus our efforts on two fronts: (i) Creating a realistic simulation environment for rapid training and evaluation and (ii) robust imitation learning method for coping with real-world imperfections. 

\subsection{Bulldozer Simulation}\label{subsec: simulation details}

Fast and efficient training is crucial for rapid evaluation and development of control algorithms. The complexity of construction site problems lies in the interaction between the vehicle and the soil. The movement of the soil due to a bulldozer's movement is not trivial and can be simulated using different techniques, each capturing a different level of detail of the real interaction \cite{10.1007/11861201_46,Sauret2014BulldozingOG,developing_simple}. 
We created a physically realistic simulated environment taking these considerations into account. 

In our simulation, each sand pile is initially modeled as a multivariate gaussian distribution as follows: 
\begin{equation} \label{eq: sand pile}
    f(x, y) = \frac{V}{2\pi\sigma_x\sigma_y} * \exp{\left(-\frac{1}{2}[(\frac{x-\mu_x}{\sigma_x})^2 + (\frac{y-\mu_y}{\sigma_y})^2)]\right)} \, ,
\end{equation}
Here, $f(x,y)$ is the height of the soil at point $x,y$. The center and footprint of the sand pile are denoted by $(\mu_x, \mu_y)[cm]$ and $(\sigma_x, \sigma_y)[cm]$ respectively and the $V[cm^3]$ is the pile's volume. For example, given a volume $V$, the pile's height is reduced as its footprint grows.

\begin{figure} [t]
    \centering
    \includegraphics[width=0.98\linewidth]{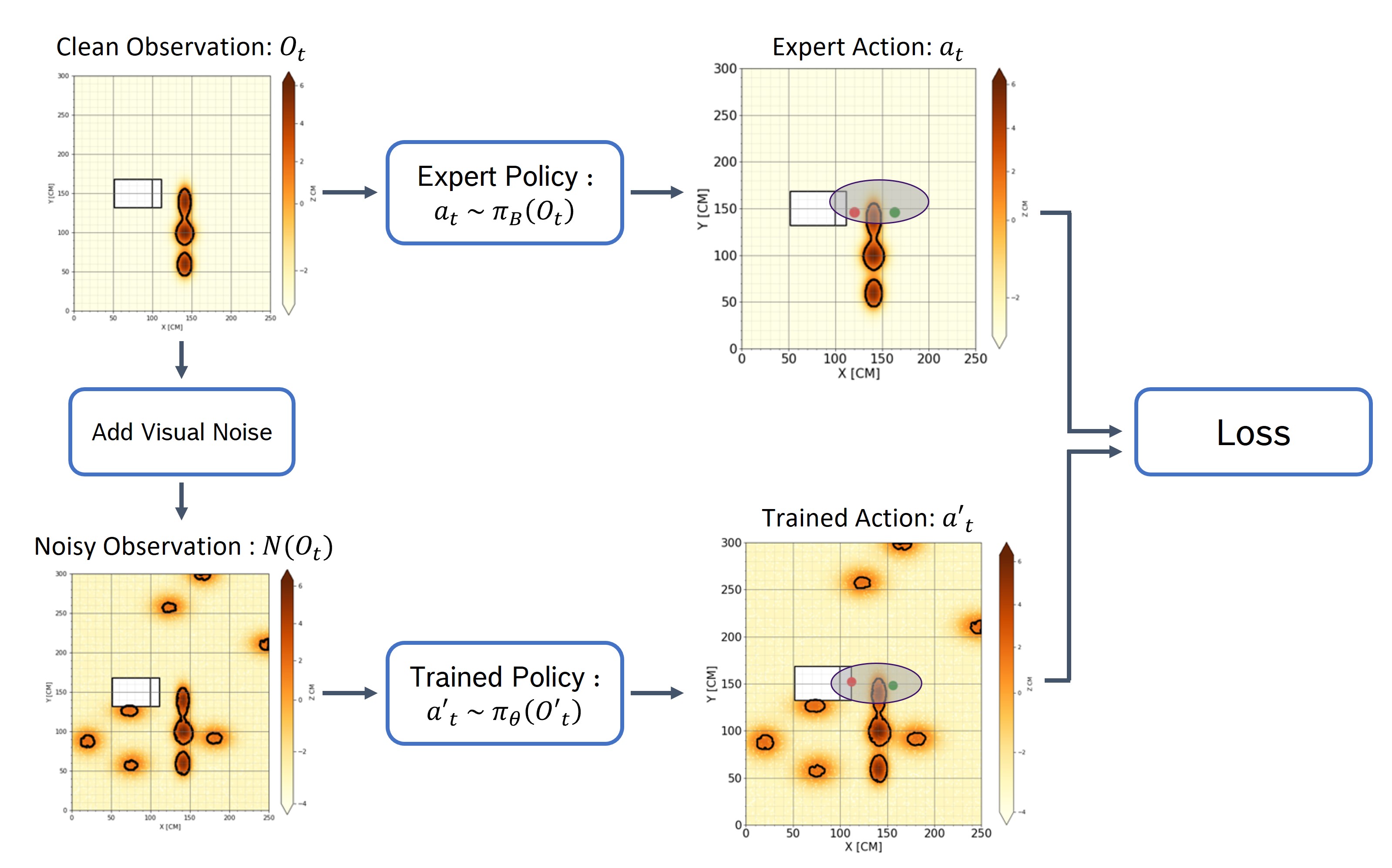}
    \caption{Illustration of the suggested privileged learning algorithm. A clean observation $o_t$ is augmented to create a noisy observation $N(o_t)$. The expert policy, i.e, \textit{baseline}, uses $o_t$ in order to produce an ideal action $a_t$ - shown as green and red waypoints. The \textit{privileged BC} agent aims to learn a robust policy $\pi_{\theta}$ that takes the augmented observation $N(o_t)$ as input and outputs an action $a'_t$ similar as possible to $a_t$.
    }
\label{fig:priviliged_learning_diagram}  
\end{figure}

Similarly to \cite{cat:spec}, we assume simple bulldozer mechanics with three possible velocity gears and a linear relation between the bulldozer's velocity and the load on the blade. This simple relation is used to calculate the total time of a specific action based on the volume of sand pushed during the performed action. In addition, we assume that the blade has a known maximal volume, which can be graded during a single leg. \cref{fig:priviliged_learning_diagram} illustrates this concept.

At its core, the bulldozer's movement is continuous, which enables the definition of low and high levels of interactions. Due to the complexity and long horizon of this task, we opted for the \textit{SnP} action-set i.e $(P,S)$, as defined in \cref{subsec: grading as mdp,subsec: baseline_algorithm}. The simulation receives an action tuple $(P,S)$, and simulates the low-level control and movement. As these actions are complex, the simulation also evaluates the time it took to perform them -- taking into consideration the volume of sand pushed and the traveled distance.

After an action is performed, the simulation returns the updated observation. To minimize the \textit{sim-to-real} gap, the observation is an \textit{ego-view} of the full heightmap, meaning a bounding box around the bulldozer rotated towards its direction of movement, as shown in \cref{fig:observation} and explained in \cref{subsec: grading as mdp}.

Although this simulation is computationally inexpensive, it takes into consideration the bulldozer's behaviour, which affects the agents ability to choose the optimal actions i.e., the change in velocity due to torque and soil-bulldozer interactions. 

\subsection{Baseline Algorithm}\label{subsec: baseline_algorithm}

Finally, after reaching $B$, the bulldozer rotates and moves towards the next starting position -- the $S$ point. This point is selected to be in front of the nearest sand pile. This logic closely mimics the behavior of a human driver\footnote{See example video in \href{https://www.youtube.com/watch?v=6dVZhFZYofY}{youtube.com/watch?v=6dVZhFZYofY}.}, efficiently grading an entire area, and takes into consideration the physical limitations of the bulldozer's movement. 

\subsection{Privileged Behavioral Cloning}\label{subsec: priv bc}

The motivation to use privileged learning techniques is the fact that the \textit{baseline} operates very well in a clean simulation environment i.e., without any noise or inaccuracies.
In the privileged learning setting, a simulator can be quite beneficial as it allows the usage of otherwise unavailable information. Specifically, a simulation can generate both clean and noisy observations, which are required for this method, where the former is unavailable in the real-world. 

The privileged behavioral cloning (PBC) technique is presented in \cref{fig:priviliged_learning_diagram}. In PBC, the agent is initialized at a random initial state, then the \textit{baseline} solves the problem while presented with noiseless observations. Both clean observations $o$ and expert actions $a$ are recorded for future use. Next, the clean observations are augmented to resemble real-world inaccuracies, denoted by $N(o)$, and presented to the agent for behavior cloning together with  the expert's actions. 


As the agent is given a more realistic noisy observation, it is required to perform robust feature extraction and learn the correct actions, thus distilling the information in the heightmap. In our case, the agent is provided an observation with multiple real-world inaccuracies. These include low-volume piles that are spread randomly around the target area, depth measurement noise and occlusions. 

Specifically, in PBC, the agent optimizes the following loss, adapted from \cref{subsec:imitation}:
\begin{equation} \label{eq: privileged loss}
    L(\theta; D) = \underset{(o, a) \sim D }{\mathbb{E}} \Bigl[ l\bigl( \pi_\theta (N(o)), a\bigr)\Bigr] ,
\end{equation}

Here, $D$ is the training dataset generated by an expert demonstrator presented with clean observations. $D$ contains corresponding actions $a$, and clean observations $o$. $N(o)$ is the augmented observations, $\pi_{\theta}$ is the agent's learned policy and $l(\cdot)$ is a cross entropy metric.



\subsection{Scaled Prototype Environment}\label{subsec: prototype dozer}
While the simulated environment enables rapid training and testing, it is not physically precise. For this reason, we built a $1:9$ scaled prototype environment, which includes a $250 \times 250cm$ sandbox and a scaled bulldozer prototype $60 \times 40cm$ in size in order to mimic the real-world inaccuracies. See \cref{fig:sim2real_all} for images of our setup. A RGBD camera, which relies on stereo matching \cite{MVG}, is mounted above the sandbox at a distance of $2.5$ meters and provides a realistic dense heightmap of the area. In addition, the camera is used to localize the prototype bulldozer within the target area using an ArUco marker \cite{romero2018speeded}, which is mounted on top of the bulldozer. This marker provides both the $x,y$ location and yaw $\phi$ of the bulldozer. The images and positions from the camera are then used by our trained agent to predict the next $(P,S)$ action-set. The bulldozer then implements a low-level controller, moving according to the chosen waypoints in a closed-loop control manner.

\subsection{Experimental Details}\label{subsec: exp details}

In order to train our \textit{BC} agent, we sample random initial states in simulation and solve these tasks using the \textit{baseline}. During the \textit{baseline} execution, we record the observation-action tuple $(o_t, a_t)$. These sets are then fed to our \textit{BC} agents using the losses presented in \cref{subsec:imitation,subsec: priv bc} with  and without added augmentations, respectively. 
For our policy model, we use a ResNet-based \cite{he2016deep} end-to-end fully convolutional neural network with dilated convolutions. The input size is ($H\times W$), and the  output size is ($H\times W \times2$) – a channel for each of the actions $(P,S)$ (See \cref{fig:snp_example}). We train both agents using 200 random initial states where in each episode, we randomize the number of sand piles, shapes, locations, volumes and initial location. The training procedure is performed using batches of $128$ states for $2000$ epochs.

\begin{table*}
    \centering
    \begin{tabular}{l|cc|cc|cc}
        \multirow{2}{*}{\textbf{Algorithm}} & \multicolumn{2}{|c|}{\textbf{Continuous}} & \multicolumn{2}{|c|}{\textbf{Edge}} & \multicolumn{2}{|c}{\textbf{Initialization}} \\ \cline{2-7}
         &  Volume cleared [\%] & Time [minutes] & Volume cleared [\%] & Time [minutes] &  Volume cleared [\%] & Time [minutes] \\ \hline
         \textbf{\textit{baseline}} &  98.2 & 11.8 & 97 & 1.7 & 98.1 & 2.2 \\
         \textbf{BC} & 88.6 & 12.9 & 93.2 & 2.1 & 97.7 & 2.8 \\
         \textbf{Privileged BC} & 87.3 & 16.4 & 93.7 & 2.8 & 95.3 & 4.5
    \end{tabular}
    \caption{Results of our algorithms in \emph{simulation} using clean observations. Each result presents an average over \textbf{50} random episodes. Volume cleared is measured in \% ($\uparrow$ is better) and time in minutes ($\downarrow$ is better). Description of each scenario is presented in \cref{fig:scenario_examples}.}
    \label{tab: clean simulation results} 
\end{table*}

\section{Experiments} \label{sec:experiments}

\begin{figure} [!ht]
    \centering
    \begin{subfigure}[b]{\linewidth} 
        \includegraphics[width=0.97\linewidth]{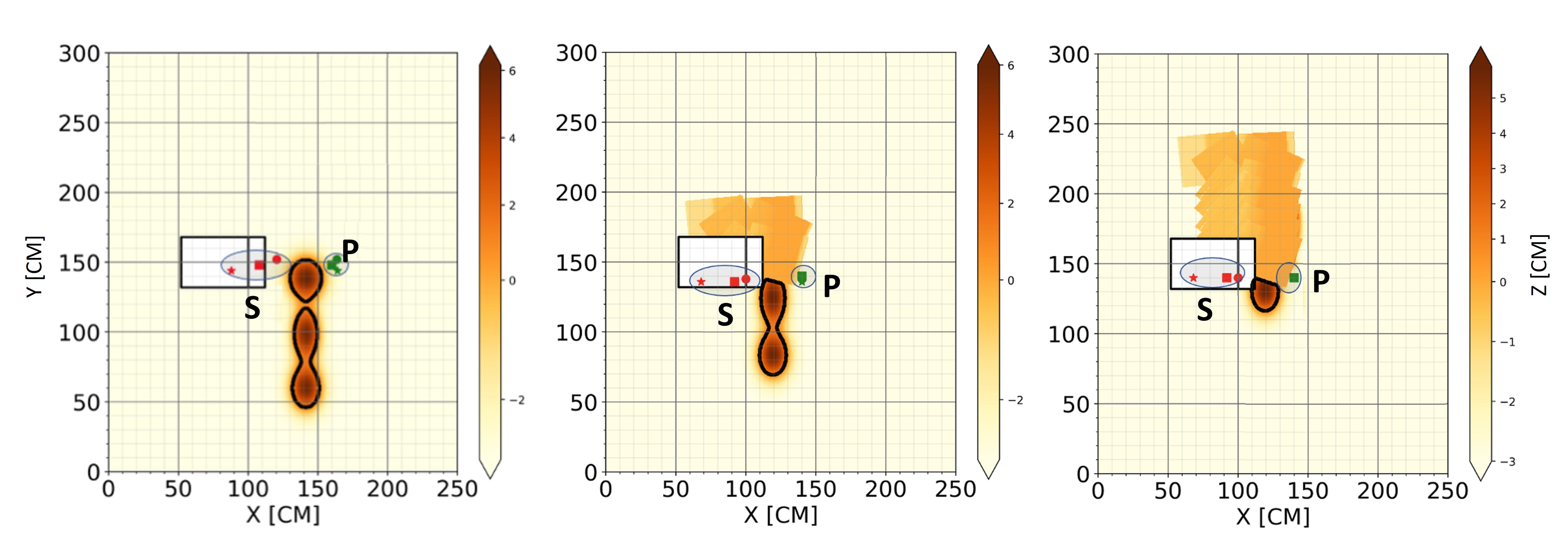}
        \caption{Clean Observations}
        \label{fig:state_action}
    \end{subfigure}
    
    \begin{subfigure}[b]{\linewidth}
        \includegraphics[width=0.99\linewidth]{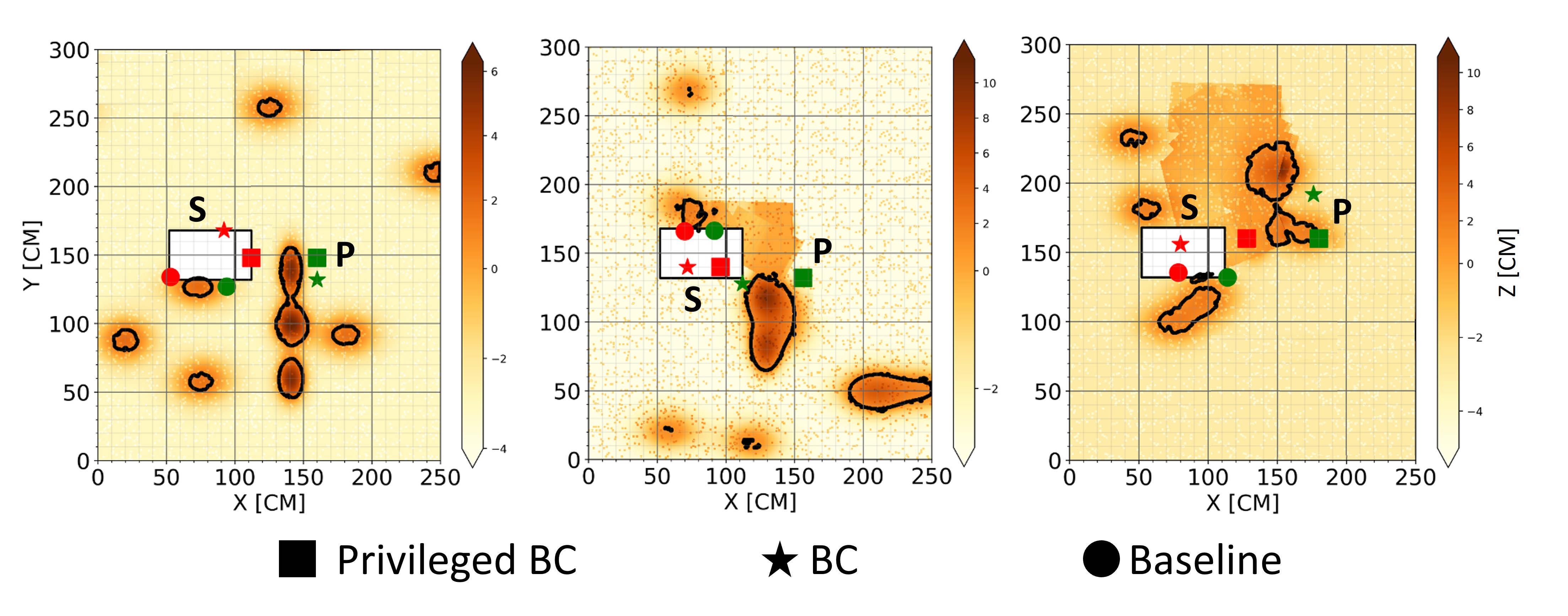}
        \caption{Noisy Observations}
        \label{fig:aug_state_action}
    \end{subfigure}
    
    \caption{Within the simulation, we compare between the various agents i.e. \textit{baseline}, \textit{BC} and \textit{privileged BC}, on both clean and noisy observations. \textbf{(a)} Evaluation on clean observations. Here, our agents are capable of predicting actions with a similar intention to that of the \textit{baseline}'s. \textbf{(b)} Evaluation on noisy observations. Here, both the \textit{baseline} and \textit{BC} agents fail to detect the sand piles as opposed to the \textit{privileged BC} agent. }
    \label{fig: simulation comparison} 
\end{figure}

We conduct rigorous experiments in both the simulated and scaled prototype environments in order to validate our assumptions. First, we prove that although the \textit{baseline} performs exceptionally well in a clean simulation, it fails catastrophically when deployed in our scaled prototype environment. Second, we train DNN based agents using BC and privileged BC and show their ability to extract valuable information and imitate the \textit{baseline}'s decision making process. 

Finally, we conduct real-world experiments showing the advantage of the privileged agent and its ability to execute the task of autonomous grading, on a real-world scaled prototype environment, despite training only in simulation.

\subsection{Simulation Results}\label{subsec:simulated_results}
Within our simulated environment, we conducted extensive experiments on the aforementioned sub-tasks, \textit{`initialization', `edge'} and \textit{`continuous'}, and methods. In each experiment, we randomize the number of sand piles, shapes, locations, volumes and initial bulldozer position. In addition, we also examined the effect of noisy observations during evaluation. Doing so allows us to fully understand the advantages and shortcomings of our proposed method. 

We train two agents, which we term \textit{BC} and \textit{privileged BC}, which were trained using clean and noisy observations respectively. In order to train our agents, we follow the methods detailed in \cref{subsec: priv bc} and \cref{subsec:imitation}. In each step of the simulation, we provide the \textit{baseline} with a clean observation $o_t$ and collect two types of data tuples $\{o_t, a_t\}$ and $\{N(o_t), a_t\}$, for the \textit{BC} and \textit{privileged BC} agents, respectively. Here $N(o_t)$ is the augmented observation described in \cref{subsec: priv bc} and $a_t$ is the action performed by the \textit{baseline}. 

\begin{table}[]
\centering
\begin{tabular}{cccc}
\multicolumn{1}{c}{}                 & \multicolumn{1}{|c|}{\textit{\textbf{baseline}}} & \multicolumn{1}{c|}{\textbf{BC}} & 
\multicolumn{1}{c}{\textbf{Privileged BC}} \\ \hline
\multicolumn{1}{c|}{\textbf{Success} $[\%]$} & \multicolumn{1}{c|}{14.3} & 
\multicolumn{1}{c|}{60}           & \multicolumn{1}{c}{\textbf{91.4}}       \\ 
\multicolumn{1}{l}{}                   & \multicolumn{1}{l}{}    & \multicolumn{1}{l}{}               & \multicolumn{1}{l}{}         
\end{tabular}
\caption{Percentage of successful actions performed on the scaled prototype environment. We examined  \textbf{35} states and the selected actions of all three agents, i.e., \textit{BC}, \textit{privileged BC} and the \textit{baseline}. An action is classified as successful if it fits the distribution of an expert on a noisy environment. Results show that \textit{privileged BC} agent outperforms other agents thus supports our hypothesis that it can generalize well to a real environment, though it was trained in a simulated one. Please refer to \cref{fig: methods comparison} for an illustration of the above.}
\label{tab:table_success_rates2}
\end{table}

Observing the behavior of the \textit{baseline} method, we conclude that it performs exceptionally well when given access to clean measurements (\cref{fig:state_action}). However, as it is designed around classic contour detection techniques, it is highly sensitive to noisy observations (See \cref{fig:aug_state_action}) where it often fails. This preliminary result inspired us to leverage the \textit{baseline} algorithm for training an imitation learning policy to generalize and cope with noisy observations. 

We can conclude that both \textit{BC} and \textit{privileged BC} agents can indeed imitate the \textit{baseline} when presented with clean observations but do not outperform it (\cref{fig:state_action} and \cref{tab: clean simulation results}). In addition, we found that the \textit{BC} agent outperforms the \textit{privileged BC} agent when presented with clean observations,   
However, as opposed to the \textit{privileged BC} paradigm, it is unable to generalize when confronted with noisy observations. This sits in line with well know theory on robustness and generalization ~\cite{wiesemann2013robust}

\begin{figure} 
    \centering

    \includegraphics[width=0.3\linewidth]{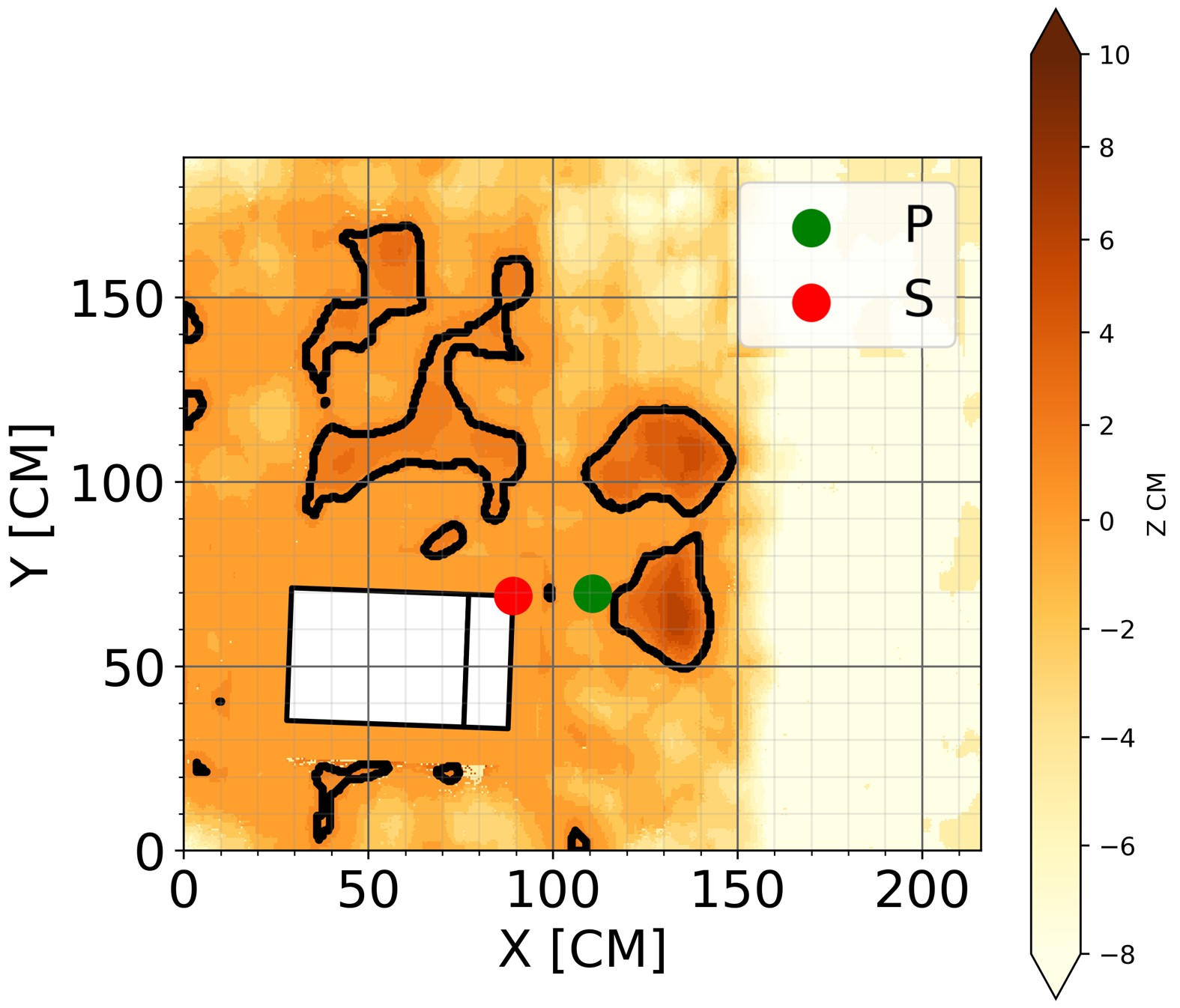}
    \hfill
    \includegraphics[width=0.3\linewidth]{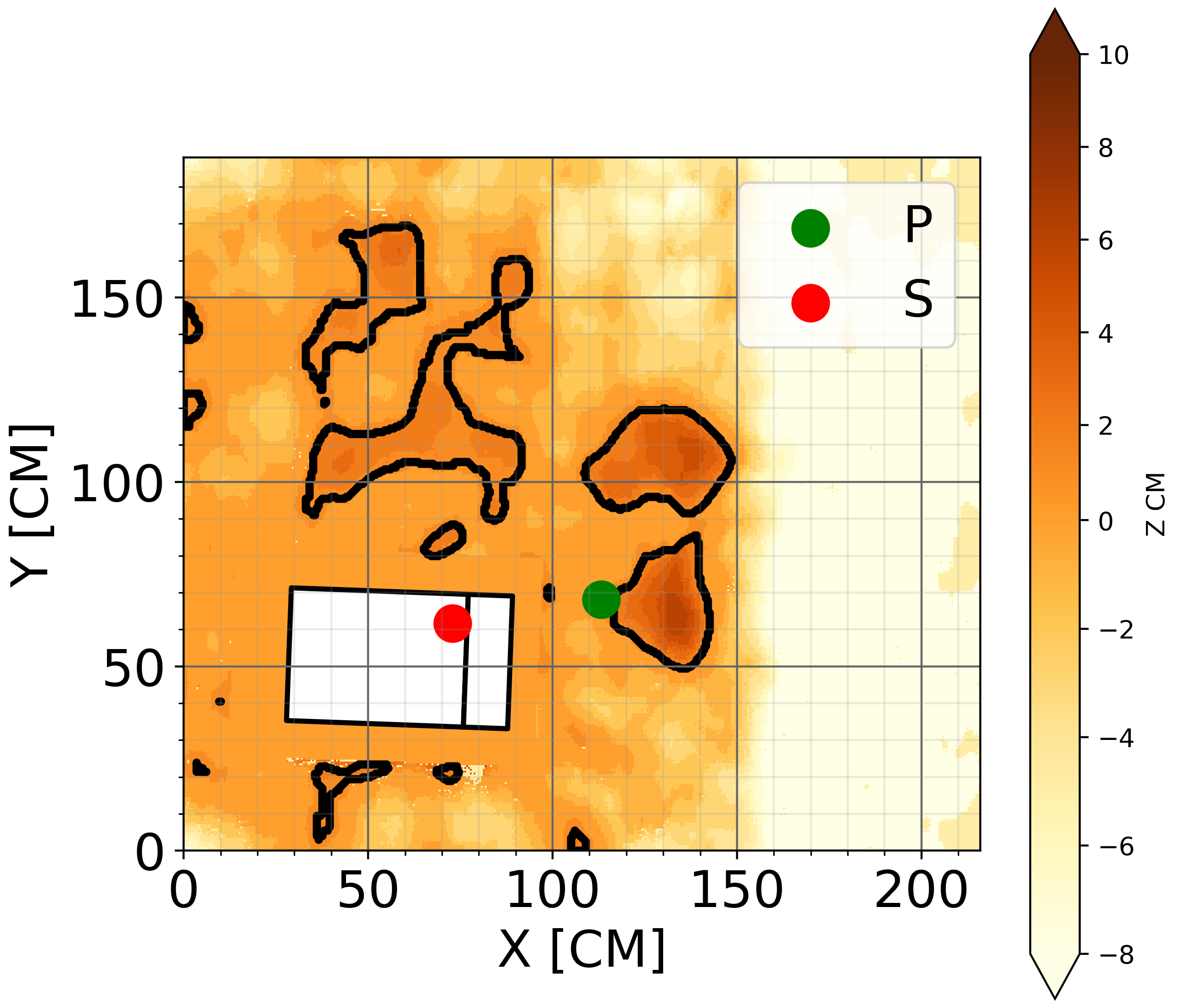}
    \hfill
    \includegraphics[width=0.3\linewidth]{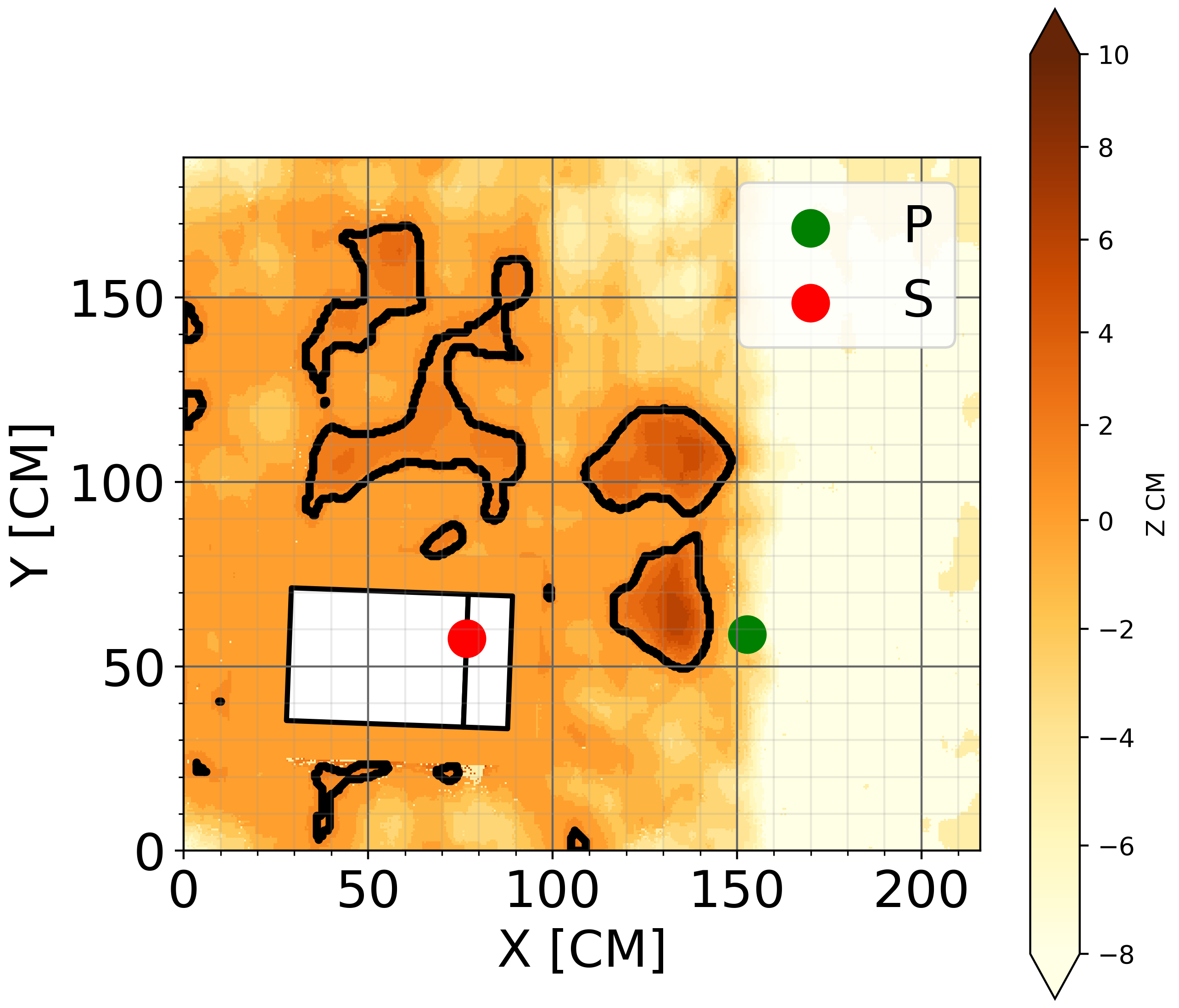}\\\vspace{1em}
    
    \includegraphics[width=0.3\linewidth]{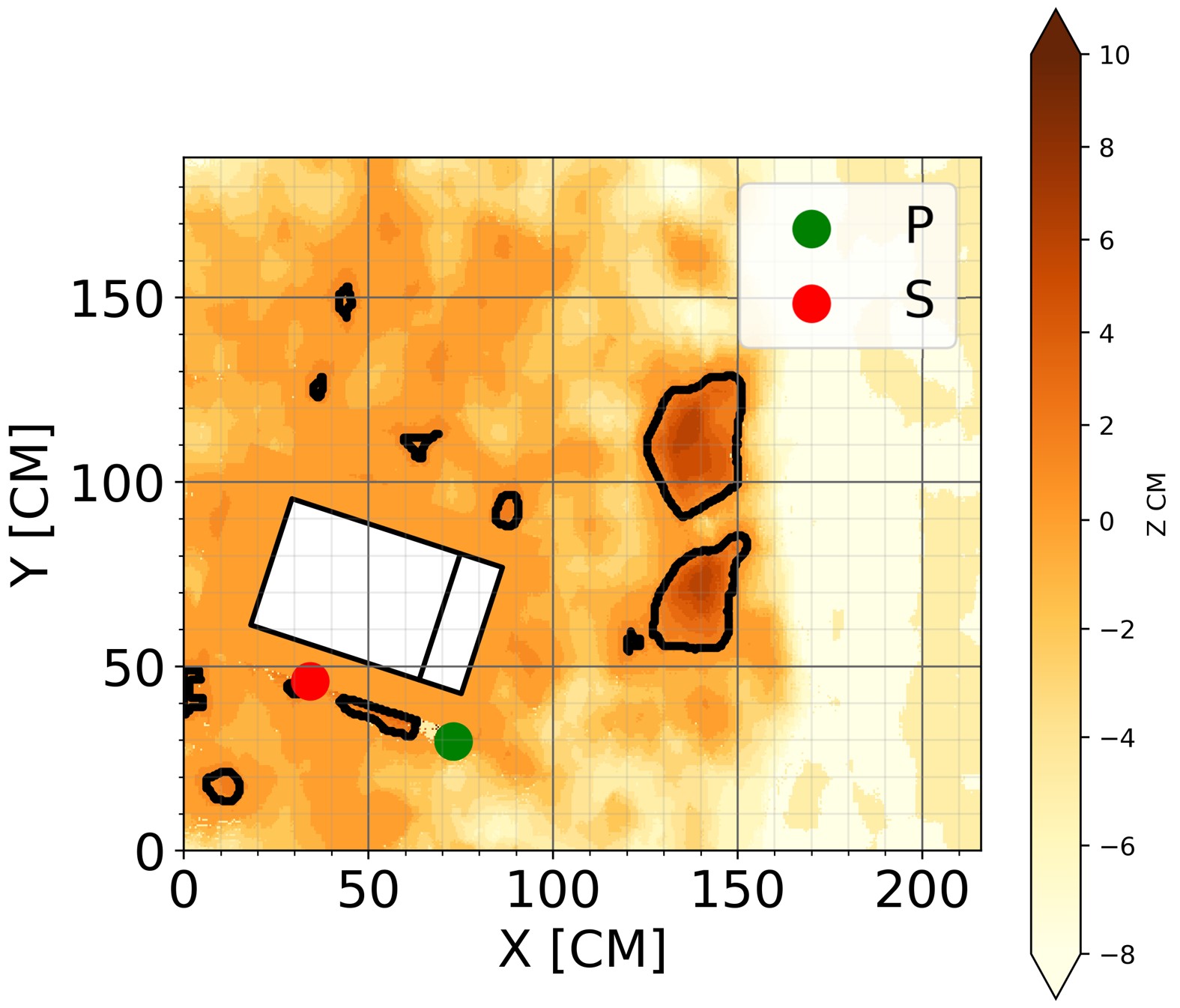}
    \hfill
    \includegraphics[width=0.3\linewidth]{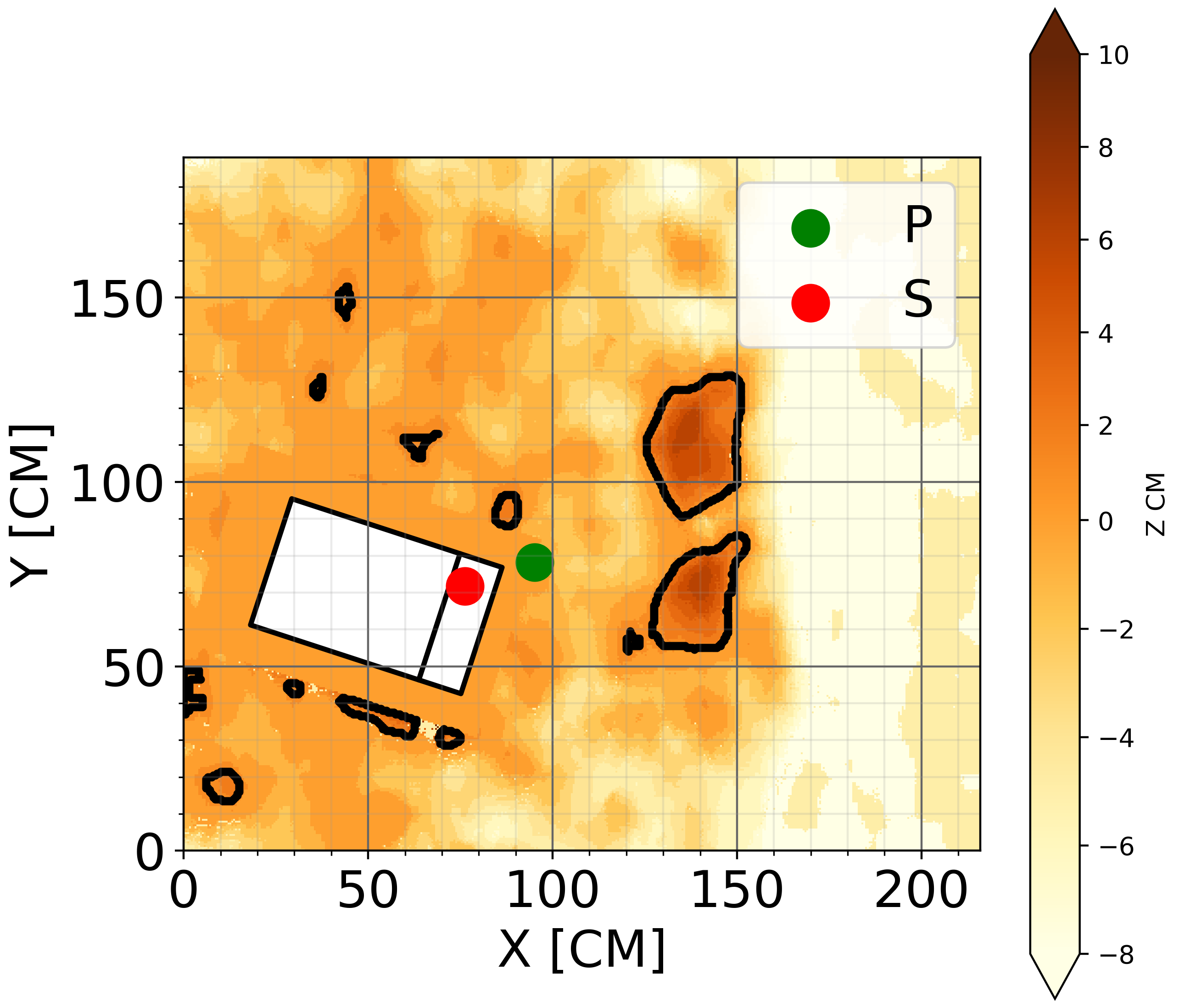}
    \hfill
    \includegraphics[width=0.3\linewidth]{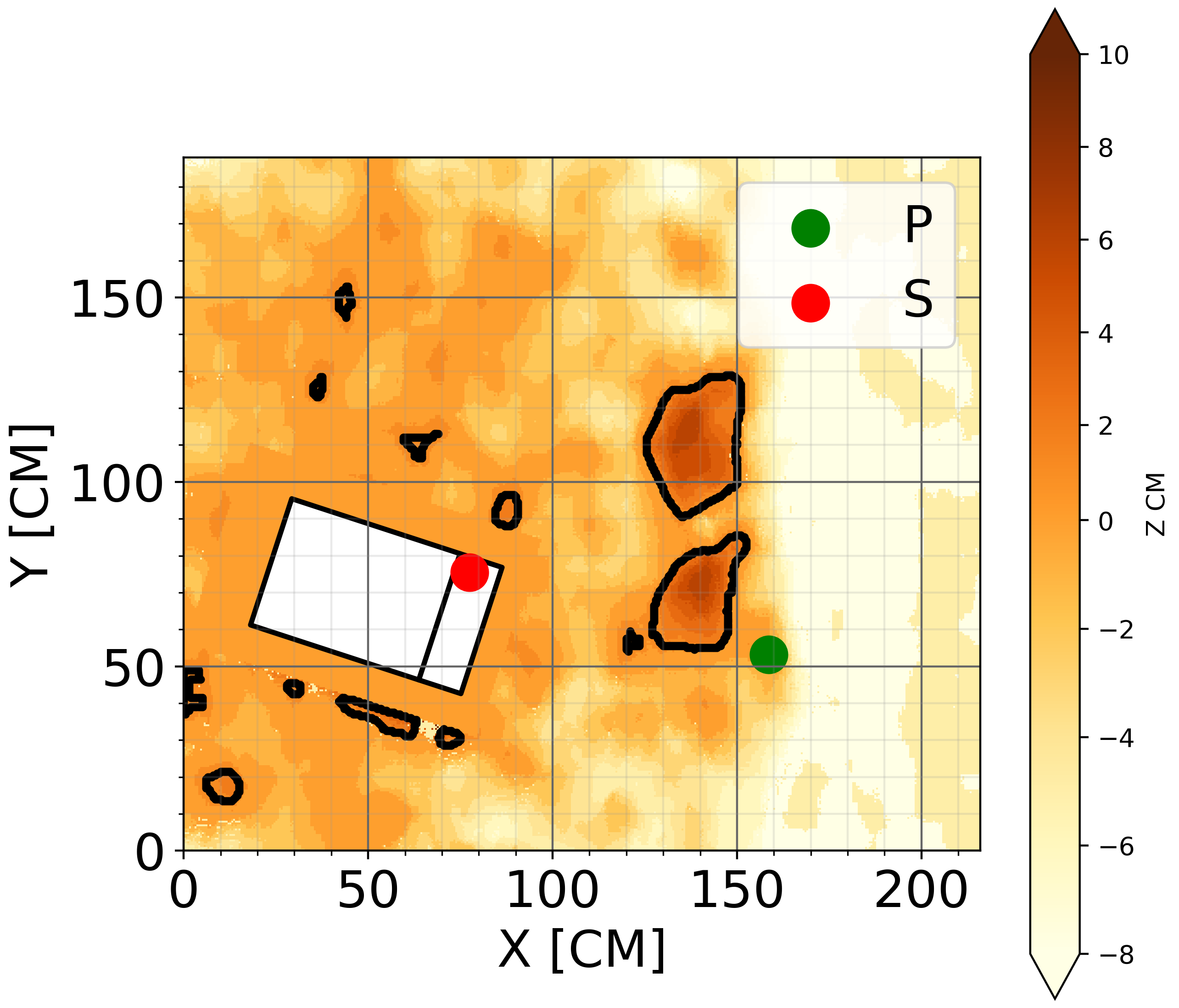}\\
    \enspace\enspace\enspace {\footnotesize (a) \textit{Baseline}} \hfill \enspace {\footnotesize (b) BC} \hfill {\footnotesize (c) Privileged BC} \enspace \,
    
    \caption{Evaluation of the various methods when placed in the real-world. The bulldozer's location is illustrated as a white element on top of the sand. The heightmap represents the sand's height at each coordinate. The contours highlight the sand piles as detected by classic detection methods and the green and red coordinates are the actions predicted by our trained agent. As can be seen, the \textit{baseline} \textbf{(a)} fails to generalize and differentiate between sand piles. The \textit{BC} agent \textbf{(b)} learns to imitate the \textit{baseline} on the clean data, and is thus prone to the same mistakes. Finally, the privileged learning paradigm \textbf{(c)} enables training a robust agent that identifies the sand piles and is capable of operating in the real-world. Please refer to \cref{tab:table_success_rates2} for quantitative results on multiple states.}
    \label{fig: methods comparison}  
\end{figure}

\begin{figure}
    \centering
    \begin{subfigure}[b]{\linewidth} 
        \includegraphics[width=0.3\linewidth]{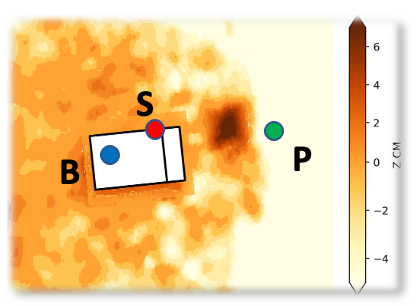}
        \hfill
        \includegraphics[width=0.3\linewidth]{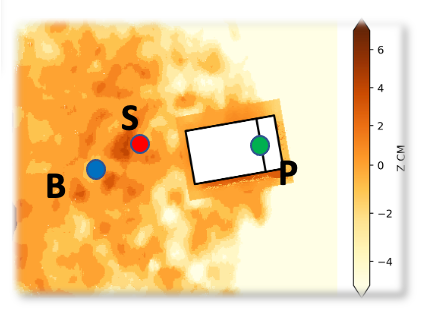}
        \hfill
        \includegraphics[width=0.3\linewidth]{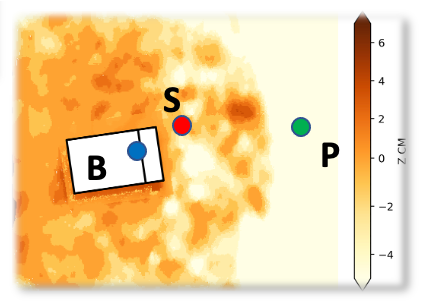}
        \caption{First $(P,S)$ set (left to right)}
        \label{fig:bc_traj_1}
    \end{subfigure}
    
    \begin{subfigure}[b]{\linewidth}
        \includegraphics[width=0.3\linewidth]{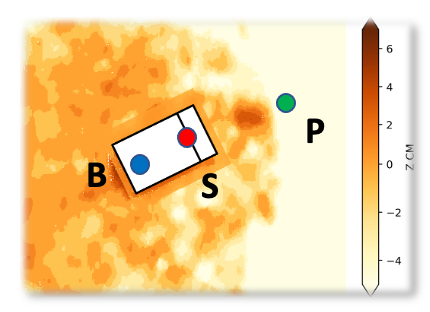}
        \hfill
        \includegraphics[width=0.3\linewidth]{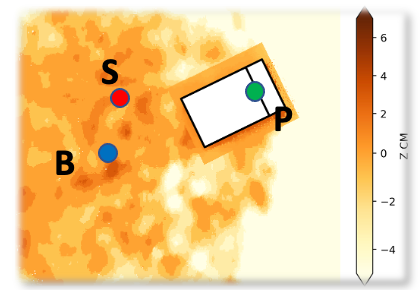}
        \hfill
        \includegraphics[width=0.3\linewidth]{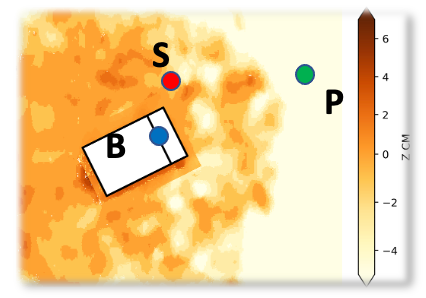}
       
        \caption{Second $(P,S)$ set (left to right)}
        \label{fig:bc_traj_2}
    \end{subfigure}
    
    \caption{A trajectory collected by the \textit{privileged BC} agent on our real-world scaled prototype environment, which includes one sand pile. Each row is a different sequence of $(P,S)$ actions chosen by the agent. The sequence is shown from left to right. The $P,S$ and $B$ points shown in \cref{fig:snp_example} are marked in red, green and blue respectively}
    \label{fig:real_traj_bc} 
\end{figure}

\subsection{Scaled Prototype Environment Results} \label{subsec:scaled_prototype_env}
Motivated by the results of \cref{subsec:simulated_results}, we continue and compare all methods on a scaled prototype environment. The prototype environment enables us to extract invaluable information regarding the true \textit{sim-to-real} gap between the simulation and an actual, real-sized, bulldozer environment, which  otherwise would be infeasible. This environment addresses two key aspects of the \textit{sim-to-real} gap. First, the underlying dynamics of our simulation are put to the test, as we cannot fully model the intricate interactions between the soil and the bulldozer. Second, the usage of a RGBD camera introduces noise and other inaccuracies into the observation space. It is important to notice that \textbf{we do not train} our agents in this environment but merely deploy the agents trained in \cref{subsec:simulated_results} and test whether they can generalize well to this environment.    

We present examples of predictions in \cref{fig: methods comparison}, and a full trajectory performed by the \textit{privileged BC} agent in \cref{fig:sim2real_all} and \cref{fig:real_traj_bc}. In addition, we present a quantitative comparison of successful actions performed on the scaled prototype environment in \cref{tab:table_success_rates2}.  These results re-validate our conclusion from \cref{subsec:simulated_results} that the \textit{baseline} algorithm under-performs when presented with real noisy observations. 

In our quantitative experiments (\cref{tab:table_success_rates2}), both \textit{BC} agents outperformed the \textit{baseline}. However, only the \textit{privileged BC} agent learned a robust feature extractor. This enabled it to solve the task with satisfying performance (over $90\%$).
and emphasizes the importance and benefit of leveraging a privileged agent within our training procedure.

\section{Conclusions} \label{sec:discussion}

In this work, we showed the importance of automating tasks in the field of construction, specifically, the grading task. We argued that data collection in the real-world is not only expensive but often infeasible. 
Our proposed simulation is beneficial for two reasons. First, it enables generation of various and diverse scenarios. Second, it allows evaluation of planning policies prior to their deployment in the real-world.
Our suggested \textit{baseline} approach, which relies on classic detection techniques, performed well in an ideal environment. However, it fails when presented with real observations that include inaccuracies. A similar behavior was observed by the \textit{BC} agent trained with clean observations.

By combining the \textit{baseline} and our simulation environment, using privileged learning, we were able to learn a robust behavior policy capable of solving the task. The \textit{privileged BC} agent was the only one able to solve a complete grading task in our scaled prototype environment.

\section{Future Work} \label{sec:future_work}

In this work, we aimed to mimic the dynamics and sensory information of a real-world bulldozer environment. Hence, future work can be divided to three  topics:

\textit{\textbf{perception}}:
In both of our environments, we used a top-view depth map, which was synthesized in simulation and acquired using a top-hanging RGBD camera on the real-world scaled prototype. However, autonomous vehicles often use on-board sensors for perception, which induce occlusions to the observation. In our case, a mounted sensor, e.g. LIDAR, will be unable to see directly behind the sand piles. We believe these occlusions can be solved using domain knowledge and \textit{sim-to-real} methods (\cref{sec:related_work}).

\textit{\textbf{dynamics}}:
Given initial and final $(P,S)$ waypoints, our low-level planner moves between the points in a straight line with periodic in-place rotations. We observed that real operators often move in a curve-like continuous path. We believe this gap can easily be solved by introducing Bézier curves \cite{farin2014curves} to the path planner. This change will improve maneuverability, both in simulation and in our real-world scaled  environment, but should not affect the overall results. 
 
\textit{\textbf{localization}}:
Our simulation and prototype did not take into account any localization errors for the bulldozer's position. While our scaled prototype environment uses an ArUco marker for localization of the bulldozer, standard autonomous systems rely on a handful of sensors in order to solve the localization task. These errors may not only affect the \textit{perception} gap but also the low-level control given the chosen actions. Future work can quantify the effect of localization errors on our overall solution.

\section{Acknowledgements}
This work is part of a joint project between \href{https://www.bosch-ai.com}{Bosch-AI} and \href{https://www.shimz.co.jp/en/}{Shimizu} aimed at making autonomous grading agents available, reliable and robust.
\bibliographystyle{IEEEtran}
\bibliography{ICRA_Dozer}

\end{document}